\documentclass[a4paper,oneside,12pt,english]{article}
\usepackage{url}
\usepackage[pdftex]{color,graphicx}
\usepackage[backref=false]{hyperref}
\hypersetup{plainpages = false,
              breaklinks=true,
              linktocpage,
              colorlinks   = true, 
              urlcolor     = blue, 
              linkcolor    = blue, 
              citecolor   = red, 
              pdfauthor={Niko Brummer},
              pdfstartpage=1, 
              pdfstartview=FitH,  
              pdfkeywords={}}

\usepackage{amsmath}%
\usepackage{amsfonts}%
\usepackage{amssymb}%
\usepackage{graphicx}

\usepackage{algorithm}
\usepackage{algpseudocode}

\usepackage{tikz}
\usetikzlibrary{bayesnet}

\usetikzlibrary{arrows,shapes,backgrounds,positioning,fit}

\def\Vmat{\mathbf{V}}

\def\ND{\mathcal{N}}

\DeclareMathOperator{\trace}{tr}

\def\traceb#1{\trace\Bigl[#1\Bigr]}

\def\expv#1#2{\left\langle#1\right\rangle_{#2}}
\def\KL#1#2{D_{KL}\bigl[#1\|#2\bigr]}

\def\R{\mathbb{R}}
\def\detm#1{\lvert#1\rvert}

\def\Cmat{\mathbf{C}}

\def\Smat{\mathbf{S}}

\def\Imat{\mathbf{I}}

\def\Xmat{\mathbf{X}}

\def\Zmat{\mathbf{Z}}
\def\Pmat{\mathbf{P}}

\def\xvec{\mathbf{x}}
\def\mvec{\mathbf{m}}

\def\muvec{\boldsymbol{\mu}}

\def\nulvec{\boldsymbol{0}}
\def\logdet#1{\log\detm{#1}}
\def\const{\text{const}}

\newcommand{\inv}[1]{%
\ifx{#1}{\Imat}           
  #1
\else
  {#1}^{-1}
\fi
}

\tikzstyle{cbox} = [rectangle,draw=blue!100,thick,align=center,rounded corners = 3pt]
\tikzstyle{lbox} = [rectangle,draw=blue!100,thick,align=left,rounded corners = 3pt]
\tikzstyle{ccircle} = [circle,draw=blue!100,thick,align=center,inner sep = 0]
\tikzstyle{ctext} = [rectangle,align=center,inner sep = 4pt]
\tikzstyle{ltext} = [rectangle,align=left,inner sep = 4pt]
\tikzstyle{solder} = [circle,draw,fill,inner sep = 0, minimum size = 3pt]

\title{How to use KL-divergence to construct conjugate priors, with well-defined non-informative limits, for the multivariate Gaussian}
\author{Niko Br\"ummer}
\date{Phonexia, September 2021}

\def\detm#1{\lvert#1\rvert}
\def\logdet#1{\log\detm{#1}}

\def\ND{\mathcal{N}}
\def\WD{\mathcal{W}}
\def\IWD{\mathcal{IW}}
\def\xvec{\mathbf{x}}
\def\Pmat{\mathbf{P}}
\def\Smat{\mathbf{S}}
\def\Cmat{\mathbf{C}}
\def\Vmat{\mathbf{V}}
\def\Xmat{\mathbf{X}}

\def\nulvec{\boldsymbol{0}}
\def\const{\text{const}}
\def\Sigmamat{\boldsymbol\Sigma}
\def\mvec{\mathbf{m}}
\def\muvec{\boldsymbol{\mu}}
\def\deltavec{\boldsymbol{\delta}}

\def\traceb#1{\trace\bigl[#1\bigr]}

\def\KL#1#2{D_\text{KL}\bigl\{#1\|#2\bigr\}}
\begin{document}

\maketitle
\begin{abstract}
The Wishart distribution is the standard conjugate prior for the precision of the multivariate Gaussian likelihood, when the mean is known---while the normal-Wishart can be used when the mean is also unknown. It is however not so obvious how to assign values to the hyperparameters of these distributions. In particular, when forming non-informative limits of these distributions, the shape (or degrees of freedom) parameter of the Wishart must be handled with care. The intuitive solution of directly interpreting the shape as a pseudocount and letting it go to zero, as proposed by some authors, violates the restrictions on the shape parameter. We show how to use the scaled KL-divergence between multivariate Gaussians as an energy function to construct Wishart and normal-Wishart conjugate priors. When used as informative priors, the salient feature of these distributions is the mode, while the KL scaling factor serves as the pseudocount. The scale factor can be taken down to the limit at zero, to form non-informative priors that do not violate the restrictions on the Wishart shape parameter. This limit is non-informative in the sense that the posterior mode is identical to the maximum likelihood estimate of the parameters of the Gaussian.                  
\end{abstract}

\section{Introduction}
Impatient readers that are familiar with the use of the Wishart distribution as a conjugate prior for Bayesian inference of the precision of a multivariate Gaussian likelihood~\cite{PRML}, may postpone reading the introduction and skip to the summary in section~\ref{sec:summary} and to the derivations in section~\ref{sec:derivations}. For a better understanding and motivation, we do however recommend reading the rest of this section---and indeed this whole note.

Consider a data set, $\Xmat=\{\xvec_i\}_{i=1}^n$, with $\xvec_i\in\R^d$, where the data is supposed to have been sampled IID from a multivariate Gaussian, $\ND(\nulvec,\Pmat^{-1})$, having zero mean and precision (inverse covariance) $\Pmat$. If the precision is unknown, we could assign to it a conjugate Wishart prior: 
\begin{align}
P(\Pmat)&=\WD(\Pmat\mid\Smat^{-1},\nu)
\intertext{and then infer $\Pmat$ via the Wishart posterior:}
P(\Pmat\mid\Xmat)&=\WD(\Pmat\mid\bar\Smat^{-1},\nu+n), &&\text{where} &
\bar\Smat &= \Smat + \sum_{i=1}^n\xvec_i\xvec_i'
\end{align}
Since the prior hyperparameter, $\Smat$, is additive in the posterior to the data scatter, $\sum_i\xvec_i\xvec_i'$, we can interpret $\Smat$ as the scatter matrix of some \emph{pseudodata}. Similarly, since the prior hyperparameter, $\nu$, is added to the count, $n$, it is tempting to interpret $\nu$ as a \emph{pseudocount}. This pseudoscatter and pseudocount interpretation is ostensibly supported by the forms of the inverses of the prior and posterior expected precisions:\footnote{Keep in mind that $\expv{\Pmat}{}^{-1}\ne\expv{\Pmat^{-1}}{}$. For example: $\expv{\Pmat^{-1}}{\text{prior}}=\frac{\Smat}{\nu-d-1}$.}
\begin{align}
\label{eq:expectations}
\expv{\Pmat}{\text{prior}}^{-1} &= \frac{\Smat}{\nu}, &
\expv{\Pmat}{\text{post}}^{-1} &= \frac{\Smat+\sum_i\xvec_i\xvec_i'}{\nu+n}
\end{align} 
where the right-hand sides are of the familiar scatter/count form of the maximum-likelihood covariance estimate. This interpretation has tempted some authors to take the prior to a non-informative limit at $\nu\to0$ and $\Smat\to\nulvec$. This is however problematic because the Wishart distribution is restricted\footnote{This restriction is analyzed more carefully in section~\ref{sec:df}.} to $\nu>d-1$. 

It should be noted that $\nu$ does have an \emph{indirect} count interpretation that derives from the definition of the Wishart: If we sample the columns of the $d$-by-$\nu$ matrix, $\Zmat$, independently from $\ND(\nulvec,\Smat^{-1})$, then $\Pmat=\Zmat\Zmat'\sim\WD(\Smat^{-1},\nu)$. But these pseudodata in $\Zmat$ are used in the prior to manufacture an example of the precision, $\Pmat$, rather than the pseudoscatter, $\Smat$. Indeed, from this Wishart definition it can be seen that if $\nu<d$, then $\Pmat$ sampled from the prior must be non-invertible and cannot function as a precision matrix.

We propose a solution to this problem by showing that we can choose the Wishart prior hyparameter to be $\nu=\alpha+d+1$, where $\alpha$ rather than $\nu$ is interpreted as the pseudocount. The proposed prior is:
\begin{align}
\label{eq:prior_summary}
\begin{split}
P(\Pmat\mid\Sigmamat,\alpha) &= \WD\bigl((\alpha\Sigmamat)^{-1},\alpha+d+1\bigr) \\
&\propto e^{-\alpha\KL{\ND(\nulvec,\Sigmamat)}{\ND(\nulvec,\Pmat^{-1})}} \\
&\propto e^{\alpha\expv{\log \ND(\xvec\mid\nulvec,\Pmat^{-1})}{\ND(\xvec\mid\nulvec,\Sigmamat)}}
\end{split}
\end{align}
where $0<\alpha\in\R$ and $\Sigmamat$ is positive definite and $D_\text{KL}$ denotes KL divergence. The KL divergence form shows that the prior mode (most likely value) is where the KL divergence vanishes, at $\Pmat=\Sigmamat^{-1}$. The last line shows the pseudocount interpretation: the prior is equivalent to augmenting the actual data set by $\alpha$ samples drawn from $\ND(0,\Sigmamat)$. The Wishart distribution remains valid for any positive real $\alpha$, so that we can now approach the limit $\alpha\to0$ to serve as non-informative prior. 

In the rest of the document, we generalize to the case of non-zero known and unknown means and provide more details, motivations and derivations.    

\section{Summary}
\label{sec:summary}
For a $d$-by-$d$, positive definite precision matrix,  $\Pmat$, we represent the Wishart distribution as:\footnote{We use the Wikipedia parametrization of the Wishart distribution: \href{https://en.wikipedia.org/wiki/Wishart_distribution}{en.wikipedia.org/wiki/Wishart-distribution}.}
\begin{align}
\label{eq:wishart}
\WD(\Pmat\mid\Vmat,\nu) &= \WD(\Pmat\mid\Smat^{-1},\nu)
\propto \detm{\Pmat}^{\frac{\nu-d-1}{2}}\, e^{-\frac12\trace(\Pmat\Smat)}
\end{align}
where we have defined $\Smat=\Vmat^{-1}$ for convenience. The parameter $\nu$ is referred to as the \emph{shape} or \emph{degrees of freedom} and $\Vmat$ is the scale matrix. We necessarily limit ourselves to Wishart distributions where the support is positive definite, in which case the shape is constrained to $\nu>d-1$ and $\Vmat$ must also be $d$-by-$d$ positive definite. 

For IID data with a multivariate normal likelihood, $\ND(\muvec,\Pmat^{-1})$ that has a known mean of $\muvec\in\R^d$, and a to-be-inferred precision, $\Pmat$, we can construct a Wishart conjugate prior as follows:
\begin{align}
\label{eq:prior1}
\begin{split}
\log P(\Pmat\mid\Sigmamat,\alpha) &= -\alpha \KL{\ND(\muvec,\Sigmamat)}{\ND(\muvec,\Pmat^{-1})} +\const \\
&= \log \WD\bigl(\Pmat\mid(\alpha\Sigmamat)^{-1},\alpha+d+1\bigr)
\end{split}
\end{align}
where $D_\text{KL}$ is KL divergence. The prior hyperparameters are $\Sigmamat$ (positive definite) and $\alpha>0$. The mode of this distribution (the prior mode) is where the KL divergence vanishes, at $\Pmat=\Sigmamat^{-1}$. If the prior is meant to be informative, the mode can be conveniently used to choose this prior hyperparameter. The KL scaling factor, $\alpha$, functions as a pseudocount, because it is additive to the data count in the parameter posterior. In the informative case, this interpretation can be used to choose $\alpha$. To obtain a non-informative prior, $\alpha$ can be taken all the way down to the limit at zero, without violating the Wishart constraint, $\nu>d-1$. At this non-informative limit, we also have that the posterior mode (MAP estimate) is identical to the maximum-likelihood (ML) estimate.   

If $\muvec$ is also unknown, we allow different means on the two sides of the KL divergence, and this still gives a conjugate prior, now in the form of the normal-Wishart:
\begin{align}
\label{eq:prior2}
\begin{split}
&\log P(\muvec,\Pmat\mid\mvec,\Sigmamat,\alpha) \\
&= -\alpha\KL{\ND(\mvec,\Sigmamat)}{\ND(\muvec,\Pmat^{-1})} +\const \\
&= \log \WD\bigl(\Pmat\mid(\alpha\Sigmamat)^{-1},\alpha+d\bigr)+\log\ND\bigl(\muvec\mid\mvec,(\alpha\Pmat)^{-1}\bigr)
\end{split}
\end{align}
The prior mode is at $(\muvec,\Pmat) = (\mvec,\Sigmamat^{-1})$, while $\alpha>0$ functions as before as the pseudocount, which can be taken down to the non-informative limit at zero. Again, at this limit, the posterior mode is identical to the ML estimate. 

It is worth noting that in the first case, the Wishart shape is $\nu=\alpha+d+1$, while in the second case it is $\nu=\alpha+d$. In the first case, for the pure Wishart prior, the prior mode is just the Wishart mode, $\Pmat=(\nu-d-1)\Vmat=\Sigmamat^{-1}$. In the second case, although the mode of the Wishart factor considered on its own is at $\frac{\alpha-1}{\alpha}\Sigmamat^{-1}$, the mode of the whole normal-Wishart product w.r.t.\ $\Pmat$ is still at $\Sigmamat^{-1}$. 

Full derivations of~\eqref{eq:prior1} and~\eqref{eq:prior2} are given in section~\ref{sec:derivations}. 

\section{Analysis}
In this section, we analyze in detail the problem with letting the shape parameter of the Wishart go to zero to form a non-informative prior. We then show that the priors defined here, \eqref{eq:prior1} and~\eqref{eq:prior2} avoid this problem.

\subsection{The Wishart distribution and its shape parameter}
\label{sec:df}
The Wishart distribution is defined as follows. A good reference that explains these details is~\cite{Dawid_81}. Let the columns of the $d$-by-$\nu$ matrix, $\Zmat$, be sampled IID from $\ND(\nulvec,\Vmat)$, then the scatter matrix, $\Zmat\Zmat'$, follows $\WD(\Vmat,\nu)$. This definition allows $\Vmat$ to be non-invertible (positive semi-definite), but in this case $\Zmat\Zmat'$ will also be non-invertible and cannot function as a precision matrix. The shape, $\nu$, can be any non-negative integer, $\nu\in\{0,1,\ldots,d,\ldots\}$. If however, $\nu\in\{0,\ldots,d-1\}$, then $\Zmat\Zmat'$ has rank $\nu<d$ (with probability 1). In particular if $\nu=0$, then the support of $\WD(\Vmat,0)$ collapses to the zero matrix. These low-rank matrices are of course only positive semi-definite and non-invertible and therefore cannot function as precision matrices. For $\nu>d-1$, the Wishart distribution can be generalized to allow non-integer, real values,\footnote{When $d=1$, this generalization is the same as the generalization from the chi-squared to the gamma distribution.} in which case (almost surely) the rank is $d$. In summary, for the Wishart to function as prior for invertible precision matrices, we need $\Vmat$ positive definite and real-valued $\nu>d-1$.  

\subsection{Limiting non-informative prior}
There are many examples in Bayesian literature where a non-informative prior is obtained by taking some distribution to an improper (unnormalizable) limit. In~\cite{PTLOS}, E.T. Jaynes supports this practice, but warns that the correct way to do this is to: 
\begin{itemize}
	\item Define everything using proper (normalizable) distributions, with one or more prior hyperparameters that can be adjusted later to the non-informative limit.
	\item Compute the posterior as a function of these adjustable parameters. 
	\item Then take the posterior to the non-informative limit. If the posterior remains proper, the result is useful.
\end{itemize}
He warns that directly taking the prior to the non-informative limit, before computing the posterior, can lead to errors. 

We can indeed apply Jaynes's recipe to the Wishart prior. But we believe the wrong way to do this is to take a real-valued $\nu$ down to the limit at zero, because the Wishart distribution does not exist for real-valued $\nu\le d-1$. Moreover, as explained above, integer-valued $\nu\le d-1$ cannot produce invertible precision matrices. Unfortunately, in an influential technical report, Tom Minka~\cite{Minka_Gaussian} does exactly this, and many authors have subsequently followed his example---see for example section 4.6.3.2 in Murphy's book~\cite{Murphy}. It should be noted that Jaynes derives a similar non-informative prior for the \emph{univariate} Gaussian in section 12.4 of~\cite{PTLOS}, but at $d=1$, we have $\nu>d-1=0$, which remains valid.          

In our prior~\eqref{eq:prior1}, we have $\nu=\alpha+d+1$ and in~\eqref{eq:prior2} we have $\nu=\alpha+d$, so that in both cases, we can let $\alpha\to0$, while $\nu>d-1$ is respected, even at $\alpha=0$. Note that at the limit, when $\alpha=0$, then both $\alpha\Sigmamat$ and $\alpha\Pmat$ vanish and the Wishart and normal factors become improper. But as long as we keep $\alpha>0$ when deriving the posterior, we are respecting Jaynes's advice and we can finally let $\alpha\to0$ in the posterior, as we show below.

\subsection{Posterior for unknown precision}
Let the data be denoted as $\Xmat=\{\xvec_i\}_{i=1}^n$, where $\xvec_i\sim\ND(\muvec,\Pmat^{-1})$ IID. When $\muvec$ is given, our prior for the precision is: 
\begin{align}
P(\Pmat\mid\Sigmamat,\alpha) 
&=\WD\bigl(\Pmat\mid(\alpha\Sigmamat)^{-1}, \alpha+d+1\bigr)
\end{align}
and the posterior is:
\begin{align}
P(\Pmat\mid\Xmat,\muvec,\Sigmamat,\alpha) 
&= \WD(\Pmat\mid\bar\Smat^{-1},n+\alpha+d+1)
\end{align}
where
\begin{align}
\bar\Smat&=\alpha\Sigmamat+\sum_{i=1}^n(\xvec_i-\muvec)(\xvec_i-\muvec)'
\end{align}
For $\alpha>0$, and $\Sigmamat$ positive definite, this posterior is always well-defined. If we have enough data, $n\ge d$ and there are no linear dependencies between data points, then we can take the posterior to the limit that makes the prior non-informative: 
\begin{align}
\lim_{\alpha\to0} P(\Pmat\mid\Xmat,\muvec,\Sigmamat,\alpha) 
&= \WD(\Pmat\mid\bar\Smat_0^{-1},n+d+1)
\end{align}
where
\begin{align}
\bar\Smat_0&=\sum_{i=1}^n(\xvec_i-\muvec)(\xvec_i-\muvec)'
\end{align}
The maximum posterior (MAP) estimate for $\Pmat$ is at the mode of the above Wishart posterior:\footnote{Be careful, the inverse of the MAP precision is \emph{not} equal to the MAP covariance: If we do a transformation of variables: $\Cmat=\Pmat^{-1}$, then we get $\Cmat\sim\IWD(\bar\Smat,n+\alpha+d+1)$, with mode $\hat\Cmat=\frac{1}{n+\alpha+2d+2}\bar\Smat\ne\hat\Pmat^{-1}=\frac{1}{n+\alpha}\bar\Smat$.}
\begin{align}
\hat\Pmat &=(n+\alpha)\bar\Smat^{-1} \\
\intertext{or}
\label{eq:mapw}
\hat\Pmat^{-1} &= \frac{\bar\Smat}{n+\alpha} = \frac{\alpha\Sigmamat+\sum_{i=1}^n(\xvec_i-\muvec)(\xvec_i-\muvec)'}{n+\alpha}
\end{align}
where it can be seen that $\alpha$ can be interpreted as a \emph{pseudocount}. In the posterior, both $\alpha$ and the prior shape, $\nu=\alpha+d+1$, are additive to the actual count $n$; but $\nu$ cannot be taken down to zero, while $\alpha$ can. We therefore propose that $\alpha$, rather than $\nu$, be interpreted as the pseudocount. At $\alpha=0$ the MAP estimate degenerates to the well-known maximum-likelihood (ML) estimate:
\begin{align}
\lim_{\alpha\to0}\hat\Pmat^{-1} &= \frac{1}{n}\sum_{i=1}^n(\xvec_i-\muvec)(\xvec_i-\muvec)'
\end{align}

\subsection{Posterior for unknown mean and precision}
In the more complex case where $\muvec$ must also be inferred, the details get more messy, but $\alpha$ retains its interpretation as pseudocount. We represent the data, $\Xmat=\{\xvec_i\}_{i=1}^n$, by its statistics: $n, \bar\xvec$ and $\tilde\Smat_0$, where
\begin{align}
\bar\xvec &= \frac1n\sum_{i=1}^n\xvec_i&&\text{and} &
\tilde\Smat_0 &= \sum_{i=1}^n(\xvec_i-\bar\xvec)(\xvec_i-\bar\xvec)'
\end{align}
The normal-Wishart prior is:
\begin{align}
 P(\muvec,\Pmat\mid\mvec,\Sigmamat,\alpha) &=\WD\bigl(\Pmat\mid(\alpha\Sigmamat)^{-1},\alpha+d\bigr)\ND\bigl(\muvec\mid\mvec,(\alpha\Pmat)^{-1}\bigr)
\end{align}
The parameter posterior can be written in identical form:\footnote{See for example: \href{https://en.wikipedia.org/wiki/Normal-Wishart_distribution}{en.wikipedia.org/wiki/Normal-Wishart-distribution}}
\begin{align}
\label{eq:normwishpost}
P(\muvec,\Pmat\mid\Xmat,\mvec,\Sigmamat,\alpha) &= \WD\bigl(\Pmat\mid(\alpha^*\Sigmamat^*)^{-1},\alpha^*+d\bigr)\,\ND\bigl(\muvec\mid\mvec^*,(\alpha^*\Pmat)^{-1}\bigr)
\end{align}
where:
\begin{align}
\begin{split}
\alpha^*\Sigmamat^* &= \alpha\Sigmamat + \tilde\Smat_0 + \frac{n\alpha}{n+\alpha}(\mvec-\bar\xvec)(\mvec-\bar\xvec)'   \\ 
\alpha^* &= \alpha + n \\
\mvec^* &= \frac{\alpha\mvec+n\bar\xvec}{\alpha+n}
\end{split}
\end{align}
So that:
\begin{align}
\Sigmamat^* &= \frac{\alpha\Sigmamat}{\alpha+n} + \frac{\tilde\Smat_0}{\alpha+n} + \frac{n\alpha}{(n+\alpha)^2}(\mvec-\bar\xvec)(\mvec-\bar\xvec)'
\end{align}
From the LHS of~\eqref{eq:prior2}, we can see the posterior mode (MAP estimate) is at:
\begin{align}
\hat\muvec &= \mvec^* &&\text{and} &
\hat\Pmat^{-1} &= \Sigmamat^*
\end{align}
Again, at the non-informative limit, the MAP estimate degenerates to the well-known ML estimate:
\begin{align}
\lim_{\alpha\to0}\hat\muvec &= \bar\xvec &&\text{and} &
\lim_{\alpha\to0}\hat\Pmat^{-1} &= \frac{1}{n}\tilde\Smat_0
\end{align}

\section{KL divergence as pseudodata}
Recalling the last line of~\eqref{eq:prior_summary}, this section gives another view of the interpretation of the KL divergence priors as pseudodata. With the prior~\eqref{eq:prior2}, we can write the parameter posterior as:
\begin{align}
\begin{split}
&\log P(\muvec,\Pmat\mid\Xmat,\mvec,\Sigmamat,\alpha) \\
&= -\alpha\KL{\ND(\mvec,\Sigmamat)}{\ND(\muvec,\Pmat^{-1})} 
+\sum_{i=1}^n \log \ND(\xvec_i\mid\muvec,\Pmat^{-1}) +\const\\
&= \alpha\expv{\log\ND(\xvec\mid\muvec,\Pmat^{-1})}{\ND(\xvec\mid\mvec,\Sigmamat)}
+\sum_{i=1}^n \log \ND(\xvec_i\mid\muvec,\Pmat^{-1}) +\const\\
\end{split}
\end{align}
This formula reinforces the interpretation that this conjugate prior effectively contributes $\alpha$ points of pseudodata sampled from $\ND(\xvec\mid\mvec,\Sigmamat)$. The case of prior~\eqref{eq:prior1} is obtained in a very similar manner, by setting $\mvec=\muvec$.

\section{Derivations} 
\label{sec:derivations}
This section gives the details of how the exponentiated KL divergences\footnote{The KL divergence between multivariate Gaussians can be found, for example at \href{https://en.wikipedia.org/wiki/Kullback_Leibler_divergence}{en.wikipedia.org/wiki/Kullback-Leibler-divergence}.} can be identified with the Wishart and normal-Wishart distributions. By using the Wishart definition~\eqref{eq:wishart}, the first prior~\eqref{eq:prior1} can be rewritten in terms of $\WD(\Pmat\mid\Smat^{-1},\nu)$:
\begin{align}
\begin{split}
\log P(\Pmat\mid\Sigmamat,\alpha) &= -\alpha \KL{\ND(\nulvec,\Sigmamat)}{\ND(\nulvec,\Pmat^{-1})} +\const \\
&= \frac{\alpha}{2}\logdet{\Pmat} -\frac12\traceb{\alpha\Sigmamat\Pmat} +\const \\
&= \frac{\nu-d-1}{2}\logdet{\Pmat} -\frac12\traceb{\Smat\Pmat} + \const\\
&= \log \WD\bigl(\Pmat\mid\alpha\Sigmamat)^{-1},\alpha+d+1\bigr)
\end{split}
\end{align}
where we have identified $\nu=\alpha+d+1$ and $\Smat=\alpha\Sigmamat$. 

The second prior~\eqref{eq:prior2} can be derived, using the short-hand, $\deltavec=\muvec-\mvec$, as:\footnote{The term $-\frac{d}{2}\log\alpha$ was absorbed into $\const$ in line 4.}
\begin{align}
\begin{split}
&\log P(\muvec,\Pmat\mid\mvec,\Sigmamat,\alpha) \\
&= -\alpha \KL{\ND(\mvec,\Sigmamat)}{\ND(\muvec,\Pmat^{-1})} +\const \\
&= \frac{\alpha}{2}\logdet{\Pmat} -\frac{\alpha}2\traceb{\Sigmamat\Pmat} -\frac{\alpha}{2}\deltavec'\Pmat\deltavec +\const \\
&= \frac{\alpha-1}{2}\logdet{\Pmat} -\frac12\traceb{\alpha\Sigmamat\Pmat} +\frac12\logdet{\alpha\Pmat} -\frac{1}{2}\deltavec'\alpha\Pmat\deltavec +\const \\
&= \frac{\nu-d-1}{2}\logdet{\Pmat} -\frac12\traceb{\Smat\Pmat} +\log\ND\bigl(\muvec\mid\mvec,(\alpha\Pmat)^{-1}\bigr) + \const\\
&= \log \WD\bigl(\Pmat\mid(\alpha\Sigmamat)^{-1},\alpha+d\bigr)+\log\ND\bigl(\muvec\mid\mvec,(\alpha\Pmat)^{-1}\bigr)
\end{split}
\end{align}
where we have identified $\nu=\alpha+d$ and $\Smat=\alpha\Sigmamat$.

\section{Note on inverse Wishart priors}
A common alternative to the Wishart as prior for the precision, $\Pmat$, is the \emph{inverse Wishart} as prior for the covariance, $\Cmat=\Pmat^{-1}$. These distributions are related as~\cite{Dawid_81}:\footnote{Murphy~\cite{Murphy} gives $\IWD(\Smat,\nu+d+1)$ in section 4.5.1, which in incorrect.}
\begin{align}
\Pmat&\sim\WD(\Smat^{-1},\nu) &&\Leftrightarrow & \Cmat&\sim\IWD(\Smat,\nu)
\end{align}
and their densities are related as:
\begin{align}
\IWD(\Cmat\mid\Smat,\nu) &= \frac{\WD(\Cmat^{-1}\mid\Smat^{-1},\nu)}{\detm{\Cmat}^{d+1}} 
\end{align}
where the denominator is the absolute value of the Jacobian determinant of the change of variables, $\Cmat\mapsto\Cmat^{-1}$. 

If we directly try to rewrite our KL priors in terms of an inverse Wishart, we find an invalid inverse Wishart, with $\nu\le d-1$ or even $\nu<0$, whenever $\alpha>0$ becomes too small. However, if we do a proper change of variables and take the Jacobian determinant into account, we \emph{can} derive valid inverse Wishart and normal-inverse-Wishart priors from the KL definitions.    

\bibliographystyle{IEEEtran}
\bibliography{kl_wishart}

\end{document}